\documentclass[11pt,3p,review,authoryear,dvipsnames]{elsarticle}

\usepackage[table,xcdraw]{xcolor}

\usepackage{amsmath,amsfonts,amssymb,amsthm}
\usepackage{float}
\usepackage{enumitem}
\usepackage{hyperref}
\hypersetup{
colorlinks=true,
linkcolor=blue,
filecolor=blue,
urlcolor=blue,
pdftitle={Overleaf Example},
pdfpagemode=FullScreen,
citecolor=blue,
}
\usepackage{cleveref}

\usepackage{listings}

\definecolor{backcolour}{rgb}{0.95,0.95,0.95}

\usepackage{xurl}
\usepackage{amsmath,bm}
\usepackage{listings}
\usepackage{subfigure}

\journal{arXiv}
\biboptions{longnamesfirst}

\begin{document}

\begin{frontmatter}

\title{\texttt{LLT}: An R package for Linear Law-based Feature Space Transformation}

\author[1,2]{Marcell T. Kurbucz\corref{cor1}}
\cortext[cor1]{Corresponding author}
\ead{kurbucz.marcell@wigner.hu}
\author[1]{P\'eter P\'osfay}
\ead{posfay.peter@wigner.hu}
\author[1]{Antal Jakov\'ac}
\ead{jakovac.antal@wigner.hu}

\address[1]{Department of Computational Sciences, Wigner Research Centre for Physics, 29-33 Konkoly-Thege Mikl\'os Street, \\ H-1121 Budapest, Hungary}
\address[2]{Institute of Data Analytics and Information Systems, Corvinus University of Budapest, 8 F\H{o}v\'am Square, \\ H-1093 Budapest, Hungary}

\begin{abstract}
The goal of the linear law-based feature space transformation (LLT) algorithm is to assist with the classification of univariate and multivariate time series. The presented R package, called \texttt{LLT}, implements this algorithm in a flexible yet user-friendly way. This package first splits the instances into training and test sets. It then utilizes time-delay embedding and spectral decomposition techniques to identify the governing patterns (called linear laws) of each input sequence (initial feature) within the training set. Finally, it applies the linear laws of the training set to transform the initial features of the test set. These steps are performed by three separate functions called \texttt{trainTest}, \texttt{trainLaw}, and \texttt{testTrans}. Their application requires a predefined data structure; however, for fast calculation, they use only built-in functions. The \texttt{LLT} R package and a sample dataset with the appropriate data structure are publicly available on GitHub.
\end{abstract}


\end{frontmatter}

\section{Introduction}
\label{S:1}

Over the past decade, time series classification (TSC) has become a crucial task of machine learning and data mining. While its growing popularity is primarily due to the rapidly increasing amount of temporal data collected by widespread sensors \citep{marussy2013success}, TSC is extensively studied across a wide variety of fields, including finance \citep{chao2019novel,kwon2019time,fons2020evaluating,feo2022financial,assis2018hybrid}, activity recognition \citep{mocanu2015factored,karim2019multivariate,wang2019deep,yang2019time,kurbucz2022facilitating,vidya2022wearable}, and biology \citep{schafer2017multivariate,rajan2018generative,elsayed2019analysis,tripto2020evaluation,bock2021machine}. Despite the large effort dedicated to this topic, it remains a challenging task due to the nature of time series data, which have large data sizes and high dimensionality and are continuously updated \citep{fu2008representing,fu2011review,zhao2017convolutional,gao2018multivariate}.

Depending on whether one or more values (features) are observed at a given time, the TSC problem can be defined as a univariate \citep{sun2019univariate,del2021auto,khan2021bidirectional} or multivariate \citep{baydogan2015learning,ruiz2021great,hao2023micos} task. In the related literature, a number of approaches have been proposed to solve both tasks, and these approaches can be divided into feature-based and distance-based methods \citep[see, e.g., ][]{susto2018time,hao2023micos}. The most commonly used feature-based methods are the discrete wavelet transform (DWT) \citep{gupta2021wavelet}, wavelet packet transform (WPT) \citep{ray2016support}, and discrete Fourier transform (DFT) \citep{kriegel2018cell}, which are used in conjunction with a classification algorithm, where dynamic time warping with the one-nearest neighbor (DTW-1NN) \citep{berndt1994using} is a typical distance-based approach.

The recently published linear law-based feature space transformation (LLT) \citep{kurbucz2022facilitating} aims to facilitate univariate and multivariate time series classification tasks by transforming the structure of the feature set (or the original time series) to make the data easier to classify. As a first step, this algorithm splits the instances into training and test sets. Then, it applies time-delay embedding and spectral decomposition techniques to identify the governing patterns (called linear laws) of each input sequence (initial feature) within the training set. Finally, it utilizes the linear laws of the training set to transform the initial features of the test set. This transformation procedure has low computational complexity and provides the opportunity to develop a learning algorithm.

This paper presents an R package called \texttt{LLT}, which is the first implementation of the LLT algorithm. This package implements LLT in a flexible yet user-friendly way while using separate functions for each computational step, which facilitates the further development of the algorithm. In addition, it does not rely on functions written by the community, which results in low computational demand. The \texttt{LLT} R package and a sample dataset with the appropriate data structure are publicly available on GitHub \citep{kurbucz2023git}. The metadata of the package is presented in Table \ref{tab:meta}.

\begin{table}[H]
\caption{Metadata of the \texttt{LLT} package}
\label{tab:meta}
\centering
\resizebox{0.9\textwidth}{!}{%
\begin{tabular}{|p{0.4\textwidth}|p{0.48\textwidth}|}
\hline
\multicolumn{1}{|c|}{\textbf{Metadata description}} & \multicolumn{1}{c|}{\textbf{Metadata contents}} \\ \hline
Current code version & v0.1.0 \\ \hline
Permanent link & \url{https://github.com/mtkurbucz/LLT} \\ \hline
Legal code license & GNU General Public License v3.0 \\ \hline
Code versioning system & Git \\ \hline
Software code languages & R \\ \hline
Operating environments and dependencies & R 4.2.2 or later. OS agnostic (Linux, OS X, MS Windows). \\ \hline
Link to developer documentation and user manual & \url{https://github.com/mtkurbucz/LLT/blob/master/README.md} \\ \hline
Support email for questions & \texttt{kurbucz.marcell@wigner.hu} \\ \hline
\end{tabular}%
}
\end{table}

The rest of this paper is organized as follows. Section \ref{S:2} presents the concept of linear laws and briefly introduces the LLT algorithm. Section \ref{S:3} and \ref{S:4} describe the structure and use of the software in detail. In Section \ref{S:5}, the application of the software is presented on an electric power consumption dataset. Finally, Section \ref{S:6} discusses the impacts of the software and provides conclusions.

\section{LLT algorithm}
\label{S:2}

This section briefly overviews the definition of linear laws and how this concept can be applied to feature space transformation. Note that the LLT algorithm is described in detail by \cite{kurbucz2022facilitating}, while derivations and proofs related to the linear laws can be found in \cite{Jakovac_time_series}.

\subsection{Linear laws of time series}
\label{S:2.1}

First, consider a generic time series $\bm{z}_t$ where $t\in\small\{1,2,...,k\small\}$ represents the time. The $l^{\text{th}}$ order ($l \in \mathbb{Z}^+$ and $l<k$) time-delay embedding \citep{takens1981dynamical} of this series is defined by:

\begin{equation}
\label{eq:A_matrix}
\bm{A}=\left(\begin{matrix}\bm{z}_{1}&\bm{z}_{2}&\cdots&\bm{z}_{l}\\\
\bm{z}_{2}&\ddots&\ddots&\vdots\\\vdots&\ddots&\ddots&\vdots\\\bm{z}_{k-l}&\cdots&\cdots&\bm{z}_{k}\\\end{matrix}\right).
\end{equation}

\noindent
Then, a symmetric $l \times l$ matrix $\bm{S}$ is generated from $\bm{A}$ as follows:

\begin{equation}
\label{eq:S_matrix}
\bm{S}=\bm{A}^\intercal\bm{A}.
\end{equation}

The term law in our case implies that we are seeking those weights that transform the values of the $\bm{S}$ matrix so that they are close to zero; that is, we seek the coefficients ($\bm{v}$) that satisfy the following equation:

\begin{equation}
\label{eq:S_law}
\bm{S} \bm{v} \approx \mathbf{0},
\end{equation}

\noindent
where $\mathbf{0}$ is a column vector containing $l$ elements of null value, $\bm{v}$ is a column vector with $l$ elements and $\bm{v} \neq \mathbf{0}$. To find the $\textbf{v}$ coefficients of Eq.~\eqref{eq:S_law}, we first perform eigendecomposition on the $\bm{S}$ matrix. Then, we select the eigenvector that is related to the smallest eigenvalue. Finally, we apply this eigenvector as $\bm{v}$ coefficients, and hereinafter, we refer to it as the linear law of $\bm{z}_t$. Note that this logic is related to principal component analysis (PCA) \citep{pearson1901liii,hotelling1933analysis}; however, in contrast to PCA, we look for components that minimize the variance of the projected data \citep[see][]{Jakovac_time_series,jakovac2022reconstruction,kurbucz2022facilitating}.

\subsection{Feature space transformation}
\label{S:2.2}

Let us consider input data as $\bm X = \{ \bm X_t \;\vert \; t\in \{1,2,\dots,k\}\}$ sets (time series), where $t$ represents the observation times. The composition of this input data can be expressed as $\bm X_t = \{ \bm x_t^{i,j} \;\vert\; i\in\{1,2,\dots,n\},~j\in \{1,2,\dots,m\}\}$, where $i$ denotes the instances and $j$ identifies the different input series (initial features) belonging to a given instance. The output $\bm y \in \{1,2,\dots,c\}$ is a vector that records the classes ($c$) of instances ($ \bm y = \{ y^{i}\in\mathbb{R}\;\vert\; i\in\{1,2,\dots,n\}\}$).

During the first step of the LLT algorithm, instances ($i$) are separated into training ($tr\in\small\{1,2,\dots,\tau\small\}$) and test ($te\in\small\{\tau+1,\tau+2,\dots,n\small\}$) sets in such a way that ensures a balanced representation of the instance classes across both sets. (For transparency, we assume that the arrangement of the instances within the dataset meets this condition for the $tr$ and $te$ sets.) We then identify the linear law (see $\bm{v}$ in Eq.~\eqref{eq:S_law}) of each input series of the training set ($\bm{x}^{1,1}_t,\bm{x}^{2,1}_t,\dots,\bm{x}^{\tau,m}_t$), thus obtaining a total of $\tau \times m$ laws (eigenvectors). These laws are grouped by input series and classes as follows: $\bm{V}^j = \{\bm{V}^j_{1},\bm{V}^j_{2},\dots,\bm{V}^j_{c}\}$, where $\bm{V}^j_{c}$ refers to the laws of the training set associated with input series $j$ and class $c$.

In the next step, $\bm{S}^{te,j}$ matrices (see Eq.~\eqref{eq:S_matrix}) are calculated from the input series of the test instance, which results in $m$ matrices per instance (one for each initial feature). We then left-multiply the $\bm{V}^j$ matrices obtained from the training set by the $\bm{S}^{te,j}$ matrices of the test set related to the same initial feature ($\bm{S}^{\tau+1,1}\bm{V}^1,\bm{S}^{\tau+1,2}\bm{V}^2,\dots,\bm{S}^{n,m}\bm{V}^m$). The laws of the $\bm{V}^j$ matrices provide an estimate of whether the $\bm{S}^{te,j}$ matrices of the test set belong to the same class as them. That is, only those columns of the $\bm{S}^{te,j}\bm{V}^j$ matrices are in proximity to the null vector with relatively small variance, for which the classes of the corresponding training and testing data match.

Finally, the dimension of the resulting matrices is reduced by a function that selects the column vectors with the smallest variance and/or absolute mean from the $\bm{S}^{te,j}\bm{V}^j$ matrices for each class. After these calculation steps, the transformed feature space of the test set has $((n-\tau) l) \times ((m c) + 1)$ dimensions with the output variable.

\begin{figure}[H]
\caption{Steps of the LLT algorithm}
\label{fig:1}
 \centering
  \includegraphics[width=0.8\textwidth]{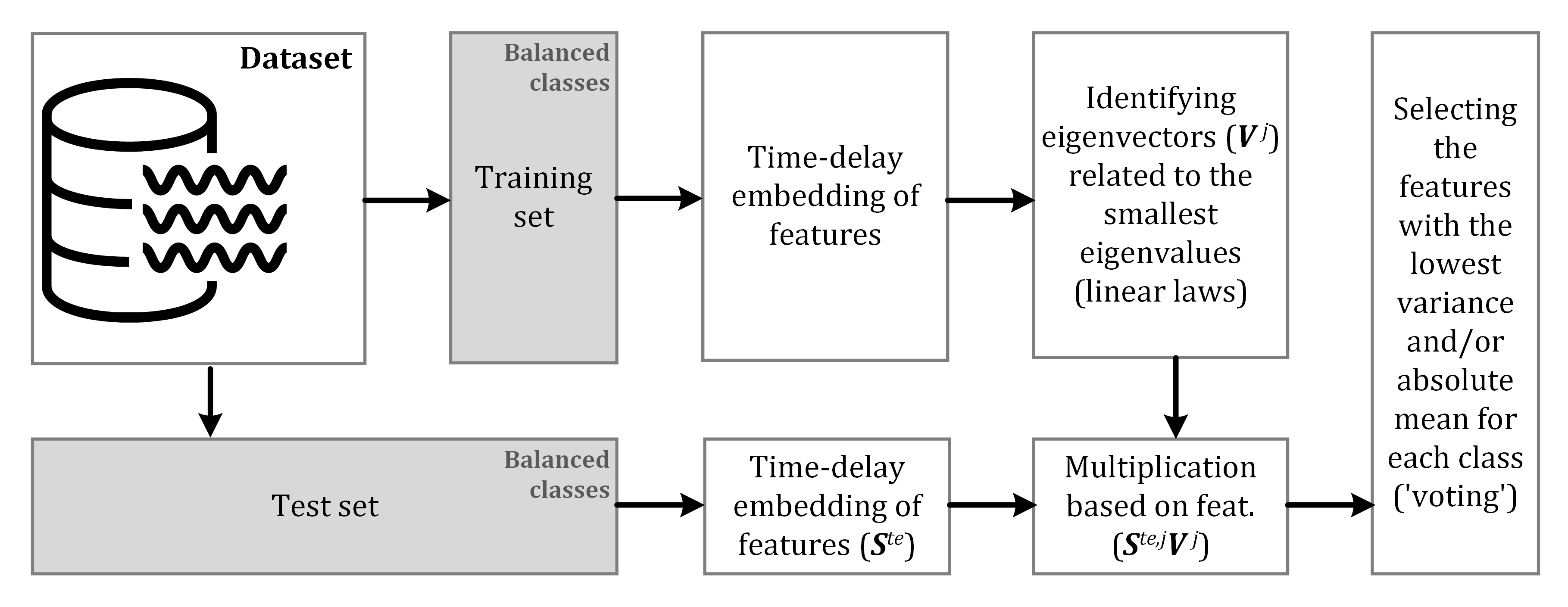}
\end{figure}

The calculation steps are illustrated in Fig. \ref{fig:1}.

\section{Software description}
\label{S:3}

The \texttt{LLT} R package is the first to implement the LLT algorithm. This package contains three main functions (\texttt{trainTest}, \texttt{trainLaw}, and \texttt{testTrans}) and two auxiliary functions (\texttt{embed} and \texttt{linlaw}). The auxiliary functions are called by the main functions, so the user does not need to use them to perform the LLT algorithm.

\vskip 0.5cm
\noindent
\underline{Description of the main functions:}
\vskip 0.3cm

\begin{sloppypar}

\begin{itemize}[noitemsep]
\item[$\bullet$]\textbf{\texttt{trainTest(path,test\_ratio,seed)}} (\textit{trainTest.R}): This function generates a two-level \textit{list} that splits the instances into training and test sets. The first level separates the training and test sets, and the second level groups the instances by class (see Fig. \ref{fig:S1}). It has two \underline{mandatory} arguments and one optional user-defined argument as follows:

\begin{itemize}[noitemsep]
\item \underline{\texttt{path}} (\textit{character}): The path to the directory that contains the instances grouped by class.
\item \underline{\texttt{test\_ratio}} (\textit{double} $\in [0,1]$): The ratio of instances in the training and test sets.
\item \texttt{seed} (\textit{integer}): The initial value of the random number seed. By default, it is not fixed.
\end{itemize}

\item[$\bullet$]\textbf{\texttt{trainLaw(path,train\_test,dim,lag)}} (\textit{trainLaw.R}): This function creates a \textit{data.frame} containing the set of laws generated from the instances of the training set. It has three \underline{mandatory} and two optional user-defined arguments as follows:

\begin{itemize}[noitemsep]
\item \underline{\texttt{path}} (\textit{character}): The path to the directory that contains the instances grouped by class.
\item \underline{\texttt{train\_test}} (\textit{list}): A two-level list that splits the instances into training and test sets. It can be generated by the \texttt{trainTest} function or defined by the user manually. Fig. \ref{fig:S1} presents an example of the appropriate structure of this object.
\item \underline{\texttt{dim}} (\textit{integer} $\in [2,k]$): It defines the row and column dimension ($l$) of the symmetric matrix $\bm{S}$. (The value $k$ is the length of the input series.)
\item \texttt{lag} (\textit{integer} $\in [1,l]$): It defines the successive row lag of the $\bm{A}$ matrix. By default, it is $1$ (see Eq.~\eqref{eq:A_matrix}). (The value $l$ is the order of the time-delayed embedding.)
\end{itemize}

\item[$\bullet$]\textbf{\texttt{testTrans(path,train\_test,train\_law,lag,select)}} (\textit{testTrans.R}): This function transforms the instances of the test set by using the LLT algorithm. It generates a \textit{data.frame} object in which columns are new features and rows are the \texttt{dim}-length time series created from the test instances and placed one below the other. It has three \underline{mandatory} and two optional user-defined arguments as follows:

\begin{itemize}[noitemsep]
\item \underline{\texttt{path}} (\textit{character}): The path to the directory that contains the instances grouped by class.
\item \underline{\texttt{train\_test}} (\textit{list}): A two-level list that splits the instances into training and test sets. It can be generated by the \texttt{trainTest} function or defined by the user manually. Fig. \ref{fig:S1} presents an example of the appropriate structure of this object.
\item \underline{\texttt{train\_law}} (\textit{data.frame}): The set of laws generated from the training instances. It can be generated by the \texttt{trainLaw} function. (For development purposes, e.g., for the creation of a learning algorithm, the user can easily modify this \textit{data.frame}.)
\item \texttt{lag} (\textit{integer} $\in [1,l]$): It defines the successive row lag of the $\bm{A}$ matrix. By default, it is $1$ (see Eq.~\eqref{eq:A_matrix}). (The value $l$ is the order of the time-delayed embedding.)
\item \texttt{select} (\textit{character} $\in \{\mathrm{"rank"},\mathrm{"var"},\mathrm{"mean"}\}$): New features are defined based on this ($f$) function (see Feature space transformation section). The "var" option selects a column vector per class and input series with the smallest variance, while the "mean" option performs this selection based on the minimum absolute mean value. The "rank" minimizes both at the same time by ranking the columns by variance and absolute mean and selecting the column with the smallest sum of ranks. All three selection criteria result in as many new features as the number of classes multiplied by the number of input series. The default value is "rank".
\end{itemize}

\end{itemize}

\vskip 0.5cm
\noindent
\underline{Description of the auxiliary functions:}
\vskip 0.3cm

\begin{itemize}[noitemsep]

\item[$\bullet$]\textbf{\texttt{embed(series,dim,lag)}} (\textit{embed.R}): This function generates the $\bm{S}$ matrix from a time series (see Eq.~\eqref{eq:S_matrix}). It has two \underline{mandatory} arguments and one optional user-defined argument as follows:

\begin{itemize}[noitemsep]
\item \underline{\texttt{series}} (\textit{numeric}): A time series in a column vector without missing values.
\item \underline{\texttt{dim}} (\textit{integer} $\in [2,k]$): It defines the row and column dimension ($l$) of the symmetric matrix $\bm{S}$. (The value $k$ is the length of the input series.)
\item \texttt{lag} (\textit{integer} $\in [1,l]$): It defines the successive row lag of the $\bm{A}$ matrix. By default, it is $1$ (see Eq.~\eqref{eq:A_matrix}). (The value $l$ is the order of the time-delayed embedding.)
\end{itemize}

\item[$\bullet$]\textbf{\texttt{linlaw(series,dim,lag)}} (\textit{linlaw.R}): By applying the \texttt{embed} function, it generates the law ($\bm{v}$) of a time series (see Eq.~\eqref{eq:S_law}). It has two \underline{mandatory} arguments and one optional user-defined argument as follows:

\begin{itemize}[noitemsep]
\item \underline{\texttt{series}} (\textit{numeric}): A time series in a column vector without missing values.
\item \underline{\texttt{dim}} (\textit{integer} $\in [2,k]$): It defines the row and column dimension ($l$) of the symmetric matrix $\bm{S}$. (The value $k$ is the length of the input series.)
\item \texttt{lag} (\textit{integer} $\in [1,l]$): It defines the successive row lag of the $\bm{A}$ matrix. By default, it is $1$ (see Eq.~\eqref{eq:A_matrix}). (The value $l$ is the order of the time-delayed embedding.)
\end{itemize}

\end{itemize}

\end{sloppypar}

The \texttt{LLT} R package and a sample dataset with the appropriate data structure are publicly available on GitHub \citep{kurbucz2023git}.

\section{Usage}
\label{S:4}

\subsection{Installation}

The \texttt{LLT} can be installed by using the \texttt{devtools} R package as follows.

\begin{lstlisting}[basicstyle=\small]
# install.packages("devtools")
# library(devtools)
devtools::install_github("mtkurbucz/LLT")
\end{lstlisting}

\subsection{Data preparation}

\begin{figure}[H]
\caption{Appropriate data structure for $2$ classes and $6$ features}
    \label{fig:2}
    \centering
\subfigure[Structure of the 'data' directory]{\includegraphics[width=.3\textwidth]{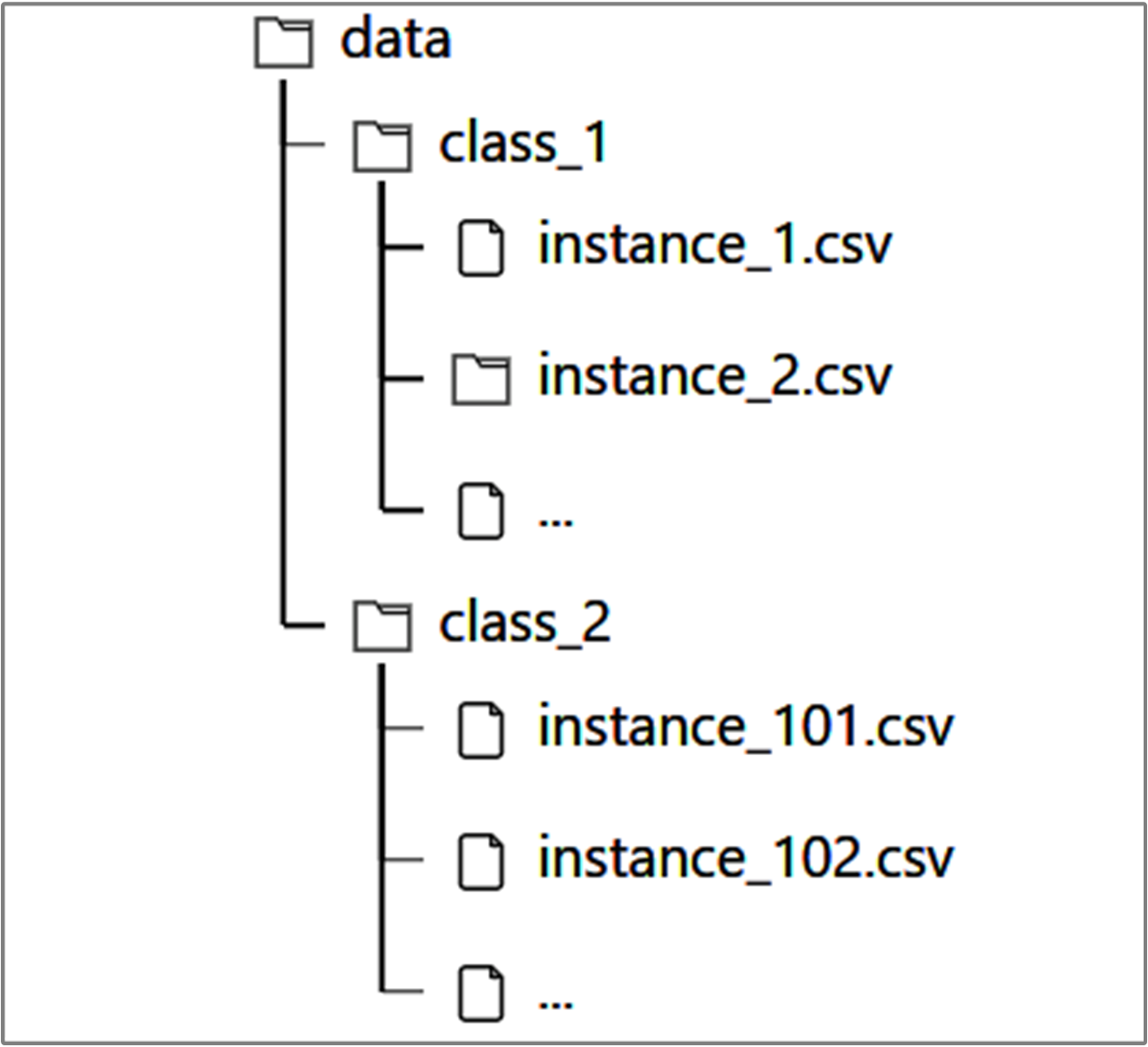}}
\hspace{1em}
\subfigure[Structure of 'instance\_1.csv' file]{\includegraphics[width=.535\textwidth]{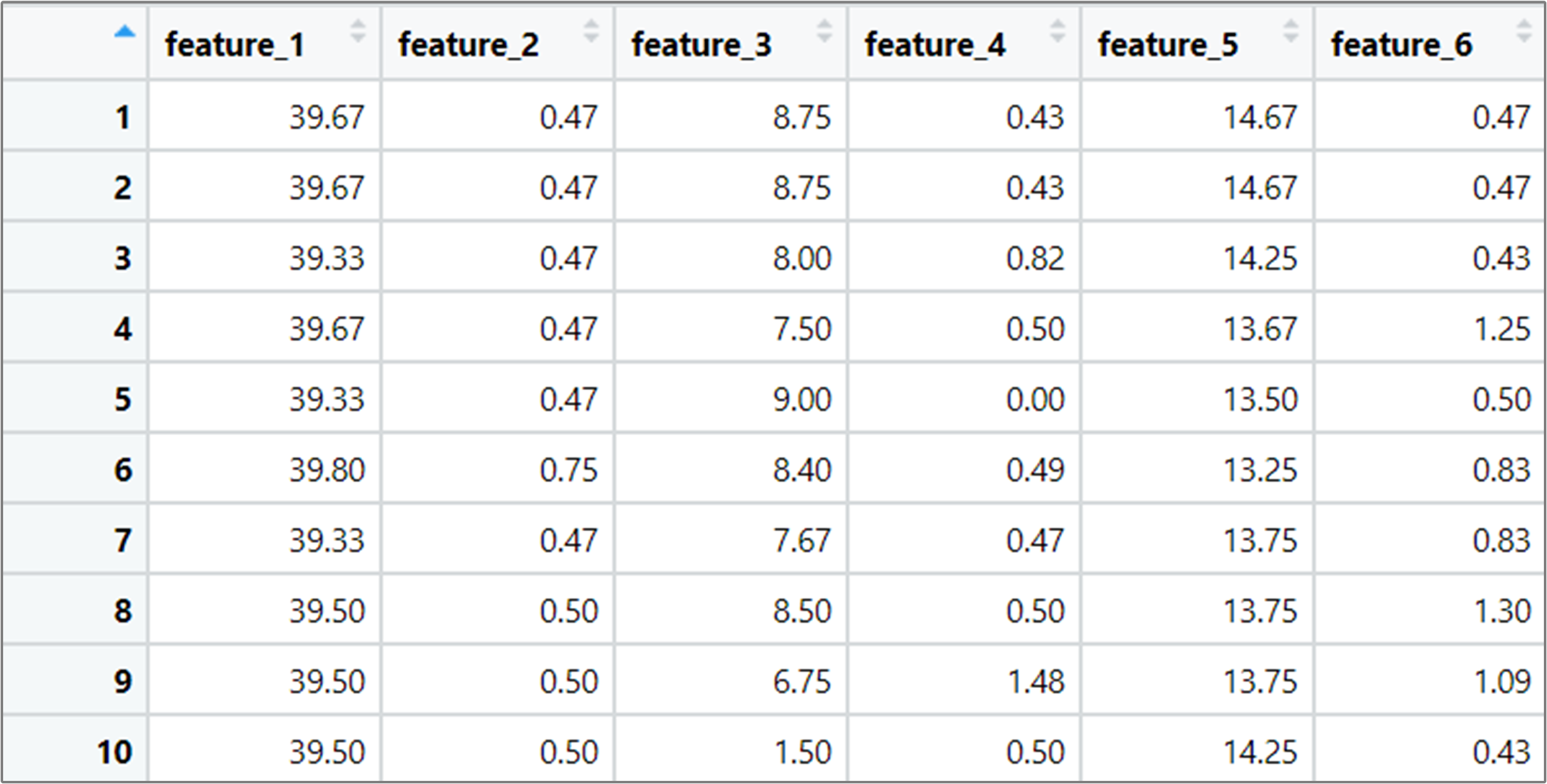}}
\end{figure}

After installation, the dataset to be transformed must be converted into a data structure in which instances are grouped by classes. Furthermore, time series features must be tab-separated column vectors with the name of the feature in the header. The appropriate data structure is presented in Fig. \ref{fig:2}.

\subsection{Data transformation}

A dataset with the appropriate structure can be transformed in the following way using the \texttt{LLT} package.

\begin{lstlisting}[basicstyle=\small]
# Loading package
library(LLT)

# Setting parameters
path <- "./data"
test_ratio <-  0.30
dim <- 9
seed <- 12345
lag <- 9
select <- "var"

# Calculation
train_test <- LLT::trainTest(path,seed,test_ratio)
train_law <- LLT::trainLaw(path,train_test,dim,lag)
result <- LLT::testTrans(path,train_test,train_law,lag,select)
\end{lstlisting}

\section{Illustrative examples}
\label{S:5}

\begin{figure}[H]
\caption{Examples of the time series belonging to each class}
    \label{fig:3}
    \centering
\subfigure[``Warm'']{\includegraphics[width=.495\textwidth]{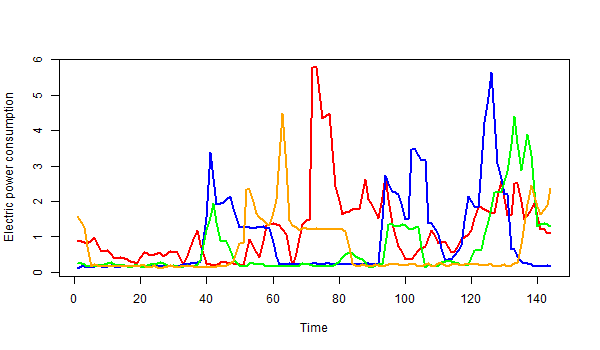}}
\subfigure[``Cold'']{\includegraphics[width=.495\textwidth]{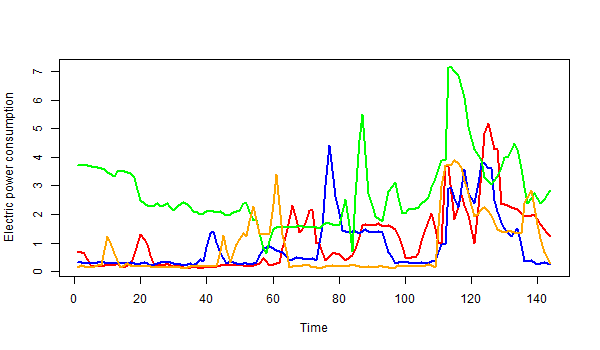}}
\end{figure}

This section presents a simple example of using the \texttt{LLT} package. In this example, we employ the PowerCons dataset collected by the Research and Development branch of Electricité de France (EDF) in Clamart (France), which is publicly available in the UCR Time Series Classification Archive \citep{UCRArchive2018}. It contains the individual household electric power consumption over the course of one year, categorized into two seasonal classes: ``Warm'' and ``Cold'', based on whether the power consumption was recorded during the warm seasons (from April to September) or the cold seasons (from October to March). Each instance in the dataset represents a day, with electric power consumption recorded at a sampling rate of ten minutes. Instances are associated with a class and comprise $144$ consecutive values. Fig. \ref{fig:3} displays examples of daily power consumption from each class.

\begin{figure}[H]
\caption{Histogram of accuracies}
\label{fig:4}
 \centering
  \includegraphics[width=0.495\textwidth]{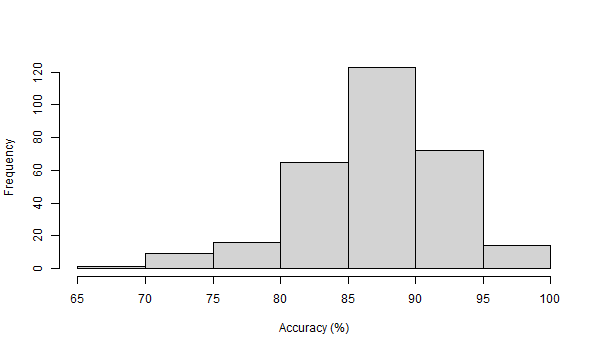}
\end{figure}

Before the transformation, we merged the training and test sets of instances that were previously separated by the authors. Then, we repeated the transformation $300$ times based on the \texttt{dim = 5} and \texttt{test\_ratio = 0.1} parameter setting. After each transformation, we calculated the mean absolute value of the resulting features for both classes and as a predicted class, we chose the class whose law resulted in a smaller absolute mean value. Based on the result of the repeated calculation procedure, we obtained an average accuracy of $87.204\%$ with a standard deviation of $5.536\%$. The histogram of accuracies achieved after each transformation is shown in Fig. \ref{fig:4}.

Note that in the case of more difficult classification tasks, it may be worthwhile to compute additional statistics (such as variance) from the new features and then apply a classification algorithm on the obtained feature space. Based on our preliminary results \citep[see, e.g., ][]{kurbucz2022facilitating}, we achieve the most accurate result with the least computational demand by combining the LLT and the k-nearest neighbor (KNN) \citep{fix1985discriminatory,cover1967nearest} algorithms.

An additional application example is provided by \cite{kurbucz2022facilitating}. In this paper, the efficiency of LLT combined with various classifiers is examined on a real-world human activity recognition (HAR) dataset called the Activity Recognition system based on Multisensor data fusion (AReM) \citep{palumbo2016human}. According to the results, LLT vastly increased the accuracy of traditional classifiers, which outperformed state-of-the-art methods after the proposed feature space transformation.

\section{Impact and conclusion}
\label{S:6}

The goal of the linear law-based feature space transformation (LLT) algorithm is to assist with the classification of univariate and multivariate time series. The presented R package, called \texttt{LLT}, implements this algorithm in a flexible yet user-friendly way. This package first splits the instances into training and test sets. It then utilizes time-delay embedding and spectral decomposition techniques to identify the governing patterns (called linear laws) of each input sequence (initial feature) within the training set. Finally, it applies the linear laws of the training set to transform the initial features of the test set. These steps are performed by three separate functions called \texttt{trainTest}, \texttt{trainLaw}, and \texttt{testTrans}. Their application requires a predefined data structure; however, for fast calculation, they use only built-in functions.

A rudimentary version of the \texttt{LLT} R package has been utilized in \cite{jakovac2022reconstruction,kurbucz2022facilitating}, and \cite{kurbucz2022linear}. Both the package and a sample dataset with the appropriate data structure are publicly available on GitHub \citep{kurbucz2023git}.

\vskip 0.5cm

\noindent
In conclusion, the value of the \texttt{LLT} R package can be summarized as follows:

\begin{itemize}[noitemsep]
\item[$\bullet$]The \texttt{LLT} package implements the linear law-based feature space transformation (LLT) algorithm in the R programming language.
\item[$\bullet$]The calculation steps are performed by separate functions, which facilitate the further development of the algorithm.
\item[$\bullet$]Despite the flexibility of the package, its functions have been designed in a user-friendly way and require only the most important parameters.
\item[$\bullet$]To maintain low computational requirements, the \texttt{LLT} package only uses built-in functions.
\end{itemize}

\section*{Data availability}

The PowerCons dataset was collected by the Research and Development branch of Electricité de France (EDF) in Clamart (France). It is publicly available in the UCR Time Series Classification Archive \citep{UCRArchive2018} at \url{http://www.timeseriesclassification.com/description.php?Dataset=PowerCons}, retrieved: 5 May 2023.

\section*{Appendix}

\renewcommand{\thefigure}{A\arabic{figure}}
\setcounter{figure}{0}

\begin{figure}[H]
\caption{Example of the structure of \texttt{train\_test} with 3 classes}
\label{fig:S1}
 \centering
  \includegraphics[width=0.6\textwidth]{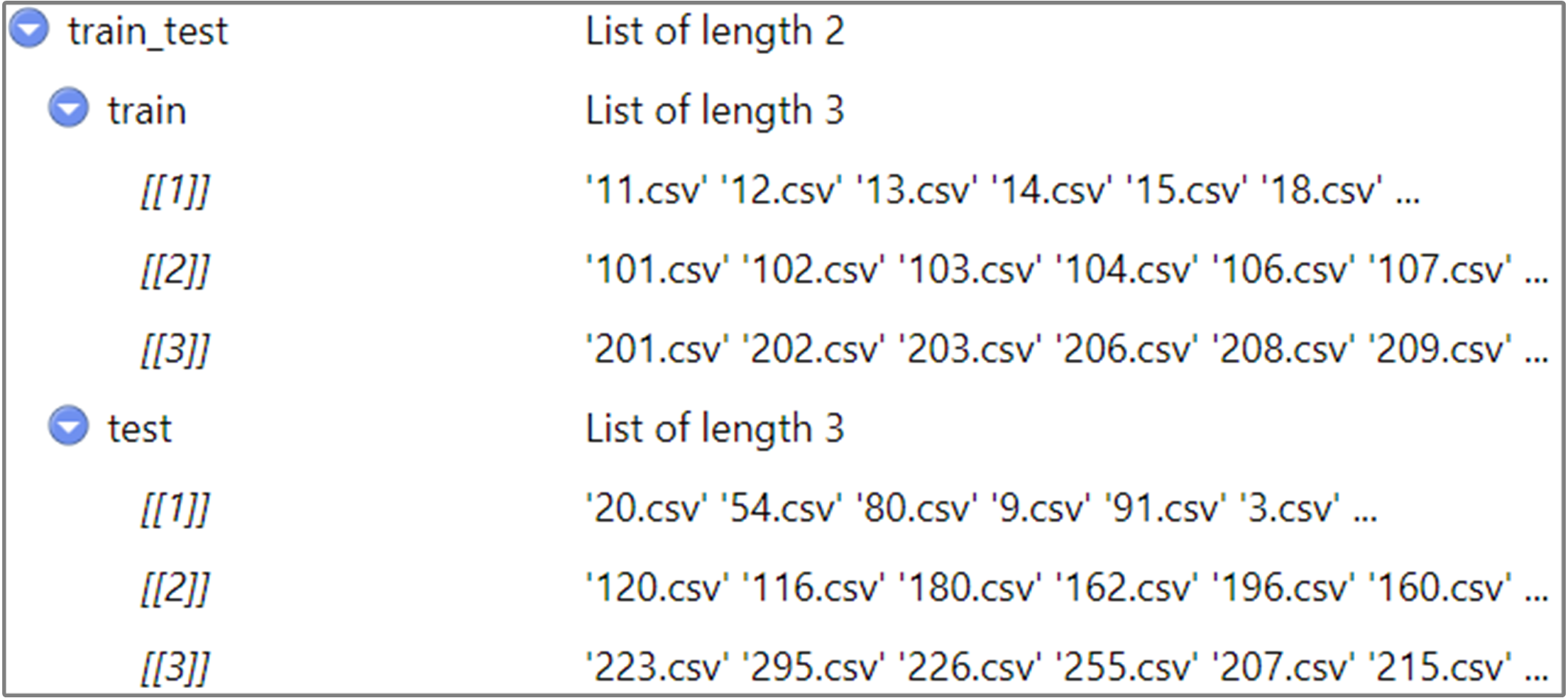}
\end{figure}

\section{Acknowledgements}

Project no. PD142593 was implemented with the support provided by the Ministry of Culture and Innovation of Hungary from the National Research, Development, and Innovation Fund, financed under the PD\_22 ``OTKA'' funding scheme. The research was supported by the Ministry of Innovation and Technology NRDI Office within the framework of the MILAB Artificial Intelligence National Laboratory Program. A.J. received support from the Hungarian Scientific Research Fund (OTKA/NRDI Office) under contract number K123815.

\bibliography{refs}

\begin{thebibliography}{43}
\expandafter\ifx\csname natexlab\endcsname\relax\def\natexlab#1{#1}\fi
\providecommand{\url}[1]{\texttt{#1}}
\providecommand{\href}[2]{#2}
\providecommand{\path}[1]{#1}
\providecommand{\DOIprefix}{doi:}
\providecommand{\ArXivprefix}{arXiv:}
\providecommand{\URLprefix}{URL: }
\providecommand{\Pubmedprefix}{pmid:}
\providecommand{\doi}[1]{\href{http://dx.doi.org/#1}{\path{#1}}}
\providecommand{\Pubmed}[1]{\href{pmid:#1}{\path{#1}}}
\providecommand{\bibinfo}[2]{#2}
\ifx\xfnm\relax \def\xfnm[#1]{\unskip,\space#1}\fi
\bibitem[{Assis et~al.(2018)Assis, Machado, Pereira \&
  Carrano}]{assis2018hybrid}
\bibinfo{author}{Assis, C.~A.}, \bibinfo{author}{Machado, E.~J.},
  \bibinfo{author}{Pereira, A.~C.}, \& \bibinfo{author}{Carrano, E.~G.}
  (\bibinfo{year}{2018}).
\newblock \bibinfo{title}{Hybrid deep learning approach for financial time
  series classification}.
\newblock {\it \bibinfo{journal}{Revista Brasileira de Computa{\c{c}}{\~a}o
  Aplicada}\/},  {\it \bibinfo{volume}{10}\/}, \bibinfo{pages}{54--63}.
\bibitem[{Baydogan \& Runger(2015)}]{baydogan2015learning}
\bibinfo{author}{Baydogan, M.~G.}, \& \bibinfo{author}{Runger, G.}
  (\bibinfo{year}{2015}).
\newblock \bibinfo{title}{Learning a symbolic representation for multivariate
  time series classification}.
\newblock {\it \bibinfo{journal}{Data Mining and Knowledge Discovery}\/},  {\it
  \bibinfo{volume}{29}\/}, \bibinfo{pages}{400--422}.
\bibitem[{Berndt \& Clifford(1994)}]{berndt1994using}
\bibinfo{author}{Berndt, D.~J.}, \& \bibinfo{author}{Clifford, J.}
  (\bibinfo{year}{1994}).
\newblock \bibinfo{title}{Using dynamic time warping to find patterns in time
  series.}
\newblock In {\it \bibinfo{booktitle}{KDD workshop}\/} (pp.
  \bibinfo{pages}{359--370}).
\newblock \bibinfo{organization}{Seattle, WA, USA:} volume \bibinfo{volume}{10,
  no. 16}.
\bibitem[{Bock et~al.(2021)Bock, Moor, Jutzeler \& Borgwardt}]{bock2021machine}
\bibinfo{author}{Bock, C.}, \bibinfo{author}{Moor, M.},
  \bibinfo{author}{Jutzeler, C.~R.}, \& \bibinfo{author}{Borgwardt, K.}
  (\bibinfo{year}{2021}).
\newblock \bibinfo{title}{Machine learning for biomedical time series
  classification: from shapelets to deep learning}.
\newblock {\it \bibinfo{journal}{Artificial Neural Networks}\/},  (pp.
  \bibinfo{pages}{33--71}).
\bibitem[{del Campo et~al.(2021)del Campo, Neri, Villegas, S{\'a}nchez,
  Dom{\'\i}nguez \& Jim{\'e}nez}]{del2021auto}
\bibinfo{author}{del Campo, F.~A.}, \bibinfo{author}{Neri, M. C.~G.},
  \bibinfo{author}{Villegas, O. O.~V.}, \bibinfo{author}{S{\'a}nchez, V.
  G.~C.}, \bibinfo{author}{Dom{\'\i}nguez, H. d. J.~O.}, \&
  \bibinfo{author}{Jim{\'e}nez, V.~G.} (\bibinfo{year}{2021}).
\newblock \bibinfo{title}{Auto-adaptive multilayer perceptron for univariate
  time series classification}.
\newblock {\it \bibinfo{journal}{Expert Systems with Applications}\/},  {\it
  \bibinfo{volume}{181}\/}, \bibinfo{pages}{115147}.
\bibitem[{Chao et~al.(2019)Chao, Zhipeng \& Yuanjie}]{chao2019novel}
\bibinfo{author}{Chao, L.}, \bibinfo{author}{Zhipeng, J.}, \&
  \bibinfo{author}{Yuanjie, Z.} (\bibinfo{year}{2019}).
\newblock \bibinfo{title}{A novel reconstructed training-set svm with roulette
  cooperative coevolution for financial time series classification}.
\newblock {\it \bibinfo{journal}{Expert Systems with Applications}\/},  {\it
  \bibinfo{volume}{123}\/}, \bibinfo{pages}{283--298}.
\bibitem[{Cover \& Hart(1967)}]{cover1967nearest}
\bibinfo{author}{Cover, T.}, \& \bibinfo{author}{Hart, P.}
  (\bibinfo{year}{1967}).
\newblock \bibinfo{title}{Nearest neighbor pattern classification}.
\newblock {\it \bibinfo{journal}{IEEE transactions on information theory}\/},
  {\it \bibinfo{volume}{13}\/}, \bibinfo{pages}{21--27}.
\bibitem[{Dau et~al.(2018)Dau, Keogh, Kamgar, Yeh, Zhu, Gharghabi,
  Ratanamahatana, Yanping, Hu, Begum, Bagnall, Mueen, Batista \&
  Hexagon-ML}]{UCRArchive2018}
\bibinfo{author}{Dau, H.~A.}, \bibinfo{author}{Keogh, E.},
  \bibinfo{author}{Kamgar, K.}, \bibinfo{author}{Yeh, C.-C.~M.},
  \bibinfo{author}{Zhu, Y.}, \bibinfo{author}{Gharghabi, S.},
  \bibinfo{author}{Ratanamahatana, C.~A.}, \bibinfo{author}{Yanping},
  \bibinfo{author}{Hu, B.}, \bibinfo{author}{Begum, N.},
  \bibinfo{author}{Bagnall, A.}, \bibinfo{author}{Mueen, A.},
  \bibinfo{author}{Batista, G.}, \& \bibinfo{author}{Hexagon-ML}
  (\bibinfo{year}{2018}).
\newblock \bibinfo{title}{{The UCR Time Series Classification Archive}}.
\newblock
  \bibinfo{note}{\url{https://www.cs.ucr.edu/~eamonn/time_series_data_2018/}}.
\bibitem[{Elsayed et~al.(2019)Elsayed, Maida \& Bayoumi}]{elsayed2019analysis}
\bibinfo{author}{Elsayed, N.}, \bibinfo{author}{Maida, A.~S.}, \&
  \bibinfo{author}{Bayoumi, M.} (\bibinfo{year}{2019}).
\newblock \bibinfo{title}{An analysis of univariate and multivariate
  electrocardiography signal classification}.
\newblock In {\it \bibinfo{booktitle}{2019 18th IEEE International Conference
  On Machine Learning And Applications (ICMLA)}\/} (pp.
  \bibinfo{pages}{396--399}).
\newblock \bibinfo{organization}{IEEE}.
\bibitem[{Feo et~al.(2022)Feo, Giordano, Niglio \& Parrella}]{feo2022financial}
\bibinfo{author}{Feo, G.}, \bibinfo{author}{Giordano, F.},
  \bibinfo{author}{Niglio, M.}, \& \bibinfo{author}{Parrella, M.~L.}
  (\bibinfo{year}{2022}).
\newblock \bibinfo{title}{Financial time series classification by nonparametric
  trend estimation}.
\newblock In {\it \bibinfo{booktitle}{Methods and Applications in
  Fluorescence}\/} (pp. \bibinfo{pages}{241--246}).
\newblock \bibinfo{organization}{Springer}.
\bibitem[{Fix(1985)}]{fix1985discriminatory}
\bibinfo{author}{Fix, E.} (\bibinfo{year}{1985}).
\newblock {\it \bibinfo{title}{Discriminatory analysis: nonparametric
  discrimination, consistency properties}\/} volume~\bibinfo{volume}{1}.
\newblock \bibinfo{publisher}{USAF school of Aviation Medicine}.
\bibitem[{Fons et~al.(2020)Fons, Dawson, Zeng, Keane \&
  Iosifidis}]{fons2020evaluating}
\bibinfo{author}{Fons, E.}, \bibinfo{author}{Dawson, P.},
  \bibinfo{author}{Zeng, X.-j.}, \bibinfo{author}{Keane, J.}, \&
  \bibinfo{author}{Iosifidis, A.} (\bibinfo{year}{2020}).
\newblock \bibinfo{title}{Evaluating data augmentation for financial time
  series classification}.
\newblock {\it \bibinfo{journal}{arXiv preprint arXiv:2010.15111}\/}, .
\bibitem[{Fu(2011)}]{fu2011review}
\bibinfo{author}{Fu, T.-c.} (\bibinfo{year}{2011}).
\newblock \bibinfo{title}{A review on time series data mining}.
\newblock {\it \bibinfo{journal}{Engineering Applications of Artificial
  Intelligence}\/},  {\it \bibinfo{volume}{24}\/}, \bibinfo{pages}{164--181}.
\bibitem[{Fu et~al.(2008)Fu, Chung, Luk \& Ng}]{fu2008representing}
\bibinfo{author}{Fu, T.-c.}, \bibinfo{author}{Chung, F.-l.},
  \bibinfo{author}{Luk, R.}, \& \bibinfo{author}{Ng, C.-m.}
  (\bibinfo{year}{2008}).
\newblock \bibinfo{title}{Representing financial time series based on data
  point importance}.
\newblock {\it \bibinfo{journal}{Engineering Applications of Artificial
  Intelligence}\/},  {\it \bibinfo{volume}{21}\/}, \bibinfo{pages}{277--300}.
\bibitem[{Gao et~al.(2018)Gao, Murphey \& Zhu}]{gao2018multivariate}
\bibinfo{author}{Gao, J.}, \bibinfo{author}{Murphey, Y.~L.}, \&
  \bibinfo{author}{Zhu, H.} (\bibinfo{year}{2018}).
\newblock \bibinfo{title}{Multivariate time series prediction of lane changing
  behavior using deep neural network}.
\newblock {\it \bibinfo{journal}{Applied Intelligence}\/},  {\it
  \bibinfo{volume}{48}\/}, \bibinfo{pages}{3523--3537}.
\bibitem[{Gupta et~al.(2021)Gupta, Seethalekshmi \& Datta}]{gupta2021wavelet}
\bibinfo{author}{Gupta, N.}, \bibinfo{author}{Seethalekshmi, K.}, \&
  \bibinfo{author}{Datta, S.~S.} (\bibinfo{year}{2021}).
\newblock \bibinfo{title}{Wavelet based real-time monitoring of electrical
  signals in distributed generation (dg) integrated system}.
\newblock {\it \bibinfo{journal}{Engineering Science and Technology, an
  International Journal}\/},  {\it \bibinfo{volume}{24}\/},
  \bibinfo{pages}{218--228}.
\bibitem[{Hao et~al.(2023)Hao, Wang, Alexander, Yuan \& Zhang}]{hao2023micos}
\bibinfo{author}{Hao, S.}, \bibinfo{author}{Wang, Z.},
  \bibinfo{author}{Alexander, A.~D.}, \bibinfo{author}{Yuan, J.}, \&
  \bibinfo{author}{Zhang, W.} (\bibinfo{year}{2023}).
\newblock \bibinfo{title}{Micos: Mixed supervised contrastive learning for
  multivariate time series classification}.
\newblock {\it \bibinfo{journal}{Knowledge-Based Systems}\/},  {\it
  \bibinfo{volume}{260}\/}, \bibinfo{pages}{110158}.
\bibitem[{Hotelling(1933)}]{hotelling1933analysis}
\bibinfo{author}{Hotelling, H.} (\bibinfo{year}{1933}).
\newblock \bibinfo{title}{Analysis of a complex of statistical variables into
  principal components.}
\newblock {\it \bibinfo{journal}{Journal of educational psychology}\/},  {\it
  \bibinfo{volume}{24}\/}, \bibinfo{pages}{417}.
\bibitem[{Jakov\'ac(2021)}]{Jakovac_time_series}
\bibinfo{author}{Jakov\'ac, A.} (\bibinfo{year}{2021}).
\newblock \bibinfo{title}{Time series analysis with dynamic law exploration}.
\newblock \URLprefix \url{https://arxiv.org/abs/2104.10970}.
  \DOIprefix\doi{10.48550/ARXIV.2104.10970}.
\bibitem[{Jakov{\'a}c et~al.(2022)Jakov{\'a}c, Kurbucz \&
  P{\'o}sfay}]{jakovac2022reconstruction}
\bibinfo{author}{Jakov{\'a}c, A.}, \bibinfo{author}{Kurbucz, M.~T.}, \&
  \bibinfo{author}{P{\'o}sfay, P.} (\bibinfo{year}{2022}).
\newblock \bibinfo{title}{Reconstruction of observed mechanical motions with
  artificial intelligence tools}.
\newblock {\it \bibinfo{journal}{New Journal of Physics}\/}, .
\bibitem[{Karim et~al.(2019)Karim, Majumdar, Darabi \&
  Harford}]{karim2019multivariate}
\bibinfo{author}{Karim, F.}, \bibinfo{author}{Majumdar, S.},
  \bibinfo{author}{Darabi, H.}, \& \bibinfo{author}{Harford, S.}
  (\bibinfo{year}{2019}).
\newblock \bibinfo{title}{Multivariate lstm-fcns for time series
  classification}.
\newblock {\it \bibinfo{journal}{Neural Networks}\/},  {\it
  \bibinfo{volume}{116}\/}, \bibinfo{pages}{237--245}.
\bibitem[{Khan et~al.(2021)Khan, Wang, Riaz, Elfatyany \&
  Karim}]{khan2021bidirectional}
\bibinfo{author}{Khan, M.}, \bibinfo{author}{Wang, H.}, \bibinfo{author}{Riaz,
  A.}, \bibinfo{author}{Elfatyany, A.}, \& \bibinfo{author}{Karim, S.}
  (\bibinfo{year}{2021}).
\newblock \bibinfo{title}{Bidirectional lstm-rnn-based hybrid deep learning
  frameworks for univariate time series classification}.
\newblock {\it \bibinfo{journal}{The Journal of Supercomputing}\/},  {\it
  \bibinfo{volume}{77}\/}, \bibinfo{pages}{7021--7045}.
\bibitem[{Kriegel et~al.(2018)Kriegel, K{\"o}hler, Bayat-Sarmadi, Bayerl,
  Hauser, Niesner, Luch \& Cseresnyes}]{kriegel2018cell}
\bibinfo{author}{Kriegel, F.~L.}, \bibinfo{author}{K{\"o}hler, R.},
  \bibinfo{author}{Bayat-Sarmadi, J.}, \bibinfo{author}{Bayerl, S.},
  \bibinfo{author}{Hauser, A.~E.}, \bibinfo{author}{Niesner, R.},
  \bibinfo{author}{Luch, A.}, \& \bibinfo{author}{Cseresnyes, Z.}
  (\bibinfo{year}{2018}).
\newblock \bibinfo{title}{Cell shape characterization and classification with
  discrete fourier transforms and self-organizing maps}.
\newblock {\it \bibinfo{journal}{Cytometry Part A}\/},  {\it
  \bibinfo{volume}{93}\/}, \bibinfo{pages}{323--333}.
\bibitem[{Kurbucz et~al.(2022{\natexlab{a}})Kurbucz, P{\'o}sfay \&
  Jakov{\'a}c}]{kurbucz2022facilitating}
\bibinfo{author}{Kurbucz, M.~T.}, \bibinfo{author}{P{\'o}sfay, P.}, \&
  \bibinfo{author}{Jakov{\'a}c, A.} (\bibinfo{year}{2022}{\natexlab{a}}).
\newblock \bibinfo{title}{Facilitating time series classification by linear
  law-based feature space transformation}.
\newblock {\it \bibinfo{journal}{Scientific Reports}\/},  {\it
  \bibinfo{volume}{12}\/}, \bibinfo{pages}{18026}.
\bibitem[{Kurbucz et~al.(2022{\natexlab{b}})Kurbucz, P{\'o}sfay \&
  Jakov{\'a}c}]{kurbucz2022linear}
\bibinfo{author}{Kurbucz, M.~T.}, \bibinfo{author}{P{\'o}sfay, P.}, \&
  \bibinfo{author}{Jakov{\'a}c, A.} (\bibinfo{year}{2022}{\natexlab{b}}).
\newblock \bibinfo{title}{Linear laws of markov chains with an application for
  anomaly detection in bitcoin prices}.
\newblock {\it \bibinfo{journal}{arXiv preprint arXiv:2201.09790}\/}, .
\bibitem[{Kurbucz et~al.(2023)Kurbucz, P{\'o}sfay \&
  Jakov{\'a}c}]{kurbucz2023git}
\bibinfo{author}{Kurbucz, M.~T.}, \bibinfo{author}{P{\'o}sfay, P.}, \&
  \bibinfo{author}{Jakov{\'a}c, A.} (\bibinfo{year}{2023}).
\newblock \bibinfo{title}{{LLT R package for Linear Law-based Feature Space
  Transformation}}.
\newblock \URLprefix \url{https://github.com/mtkurbucz/LLT}.
\bibitem[{Kwon et~al.(2019)Kwon, Kim, Heo, Kim \& Han}]{kwon2019time}
\bibinfo{author}{Kwon, D.-H.}, \bibinfo{author}{Kim, J.-B.},
  \bibinfo{author}{Heo, J.-S.}, \bibinfo{author}{Kim, C.-M.}, \&
  \bibinfo{author}{Han, Y.-H.} (\bibinfo{year}{2019}).
\newblock \bibinfo{title}{Time series classification of cryptocurrency price
  trend based on a recurrent lstm neural network}.
\newblock {\it \bibinfo{journal}{Journal of Information Processing Systems}\/},
   {\it \bibinfo{volume}{15}\/}, \bibinfo{pages}{694--706}.
\bibitem[{Marussy \& Buza(2013)}]{marussy2013success}
\bibinfo{author}{Marussy, K.}, \& \bibinfo{author}{Buza, K.}
  (\bibinfo{year}{2013}).
\newblock \bibinfo{title}{Success: a new approach for semi-supervised
  classification of time-series}.
\newblock In {\it \bibinfo{booktitle}{International Conference on Artificial
  Intelligence and Soft Computing}\/} (pp. \bibinfo{pages}{437--447}).
\newblock \bibinfo{organization}{Springer}.
\bibitem[{Mocanu et~al.(2015)Mocanu, Ammar, Lowet, Driessens, Liotta, Weiss \&
  Tuyls}]{mocanu2015factored}
\bibinfo{author}{Mocanu, D.~C.}, \bibinfo{author}{Ammar, H.~B.},
  \bibinfo{author}{Lowet, D.}, \bibinfo{author}{Driessens, K.},
  \bibinfo{author}{Liotta, A.}, \bibinfo{author}{Weiss, G.}, \&
  \bibinfo{author}{Tuyls, K.} (\bibinfo{year}{2015}).
\newblock \bibinfo{title}{Factored four way conditional restricted boltzmann
  machines for activity recognition}.
\newblock {\it \bibinfo{journal}{Pattern Recognition Letters}\/},  {\it
  \bibinfo{volume}{66}\/}, \bibinfo{pages}{100--108}.
\bibitem[{Palumbo et~al.(2016)Palumbo, Gallicchio, Pucci \&
  Micheli}]{palumbo2016human}
\bibinfo{author}{Palumbo, F.}, \bibinfo{author}{Gallicchio, C.},
  \bibinfo{author}{Pucci, R.}, \& \bibinfo{author}{Micheli, A.}
  (\bibinfo{year}{2016}).
\newblock \bibinfo{title}{Human activity recognition using multisensor data
  fusion based on reservoir computing}.
\newblock {\it \bibinfo{journal}{Journal of Ambient Intelligence and Smart
  Environments}\/},  {\it \bibinfo{volume}{8}\/}, \bibinfo{pages}{87--107}.
\bibitem[{Pearson(1901)}]{pearson1901liii}
\bibinfo{author}{Pearson, K.} (\bibinfo{year}{1901}).
\newblock \bibinfo{title}{Liii. on lines and planes of closest fit to systems
  of points in space}.
\newblock {\it \bibinfo{journal}{The London, Edinburgh, and Dublin
  philosophical magazine and journal of science}\/},  {\it
  \bibinfo{volume}{2}\/}, \bibinfo{pages}{559--572}.
\bibitem[{Rajan \& Thiagarajan(2018)}]{rajan2018generative}
\bibinfo{author}{Rajan, D.}, \& \bibinfo{author}{Thiagarajan, J.~J.}
  (\bibinfo{year}{2018}).
\newblock \bibinfo{title}{A generative modeling approach to limited channel ecg
  classification}.
\newblock In {\it \bibinfo{booktitle}{2018 40th Annual International Conference
  of the IEEE Engineering in Medicine and Biology Society (EMBC)}\/} (pp.
  \bibinfo{pages}{2571--2574}).
\newblock \bibinfo{organization}{IEEE}.
\bibitem[{Ray \& Mishra(2016)}]{ray2016support}
\bibinfo{author}{Ray, P.}, \& \bibinfo{author}{Mishra, D.~P.}
  (\bibinfo{year}{2016}).
\newblock \bibinfo{title}{Support vector machine based fault classification and
  location of a long transmission line}.
\newblock {\it \bibinfo{journal}{Engineering science and technology, an
  international journal}\/},  {\it \bibinfo{volume}{19}\/},
  \bibinfo{pages}{1368--1380}.
\bibitem[{Ruiz et~al.(2021)Ruiz, Flynn, Large, Middlehurst \&
  Bagnall}]{ruiz2021great}
\bibinfo{author}{Ruiz, A.~P.}, \bibinfo{author}{Flynn, M.},
  \bibinfo{author}{Large, J.}, \bibinfo{author}{Middlehurst, M.}, \&
  \bibinfo{author}{Bagnall, A.} (\bibinfo{year}{2021}).
\newblock \bibinfo{title}{The great multivariate time series classification
  bake off: a review and experimental evaluation of recent algorithmic
  advances}.
\newblock {\it \bibinfo{journal}{Data Mining and Knowledge Discovery}\/},  {\it
  \bibinfo{volume}{35}\/}, \bibinfo{pages}{401--449}.
\bibitem[{Sch{\"a}fer \& Leser(2017)}]{schafer2017multivariate}
\bibinfo{author}{Sch{\"a}fer, P.}, \& \bibinfo{author}{Leser, U.}
  (\bibinfo{year}{2017}).
\newblock \bibinfo{title}{Multivariate time series classification with weasel+
  muse}.
\newblock {\it \bibinfo{journal}{arXiv preprint arXiv:1711.11343}\/}, .
\bibitem[{Sun et~al.(2019)Sun, Yang, Liu, Chen, Rao \& Bai}]{sun2019univariate}
\bibinfo{author}{Sun, J.}, \bibinfo{author}{Yang, Y.}, \bibinfo{author}{Liu,
  Y.}, \bibinfo{author}{Chen, C.}, \bibinfo{author}{Rao, W.}, \&
  \bibinfo{author}{Bai, Y.} (\bibinfo{year}{2019}).
\newblock \bibinfo{title}{Univariate time series classification using
  information geometry}.
\newblock {\it \bibinfo{journal}{Pattern Recognition}\/},  {\it
  \bibinfo{volume}{95}\/}, \bibinfo{pages}{24--35}.
\bibitem[{Susto et~al.(2018)Susto, Cenedese \& Terzi}]{susto2018time}
\bibinfo{author}{Susto, G.~A.}, \bibinfo{author}{Cenedese, A.}, \&
  \bibinfo{author}{Terzi, M.} (\bibinfo{year}{2018}).
\newblock \bibinfo{title}{Time-series classification methods: Review and
  applications to power systems data}.
\newblock {\it \bibinfo{journal}{Big data application in power systems}\/},
  (pp. \bibinfo{pages}{179--220}).
\bibitem[{Takens(1981)}]{takens1981dynamical}
\bibinfo{author}{Takens, F.} (\bibinfo{year}{1981}).
\newblock \bibinfo{title}{Dynamical systems and turbulence, eds. rand, da \&
  young, l.-s}.
\newblock {\it \bibinfo{journal}{Lecture Notes in Mathematics}\/},  {\it
  \bibinfo{volume}{898}\/}, \bibinfo{pages}{366}.
\bibitem[{Tripto et~al.(2020)Tripto, Kabir, Bayzid \&
  Rahman}]{tripto2020evaluation}
\bibinfo{author}{Tripto, N.~I.}, \bibinfo{author}{Kabir, M.},
  \bibinfo{author}{Bayzid, M.~S.}, \& \bibinfo{author}{Rahman, A.}
  (\bibinfo{year}{2020}).
\newblock \bibinfo{title}{Evaluation of classification and forecasting methods
  on time series gene expression data}.
\newblock {\it \bibinfo{journal}{Plos one}\/},  {\it \bibinfo{volume}{15}\/},
  \bibinfo{pages}{e0241686}.
\bibitem[{Vidya \& Sasikumar(2022)}]{vidya2022wearable}
\bibinfo{author}{Vidya, B.}, \& \bibinfo{author}{Sasikumar, P.}
  (\bibinfo{year}{2022}).
\newblock \bibinfo{title}{Wearable multi-sensor data fusion approach for human
  activity recognition using machine learning algorithms}.
\newblock {\it \bibinfo{journal}{Sensors and Actuators A: Physical}\/},  {\it
  \bibinfo{volume}{341}\/}, \bibinfo{pages}{113557}.
\bibitem[{Wang et~al.(2019)Wang, Chen, Hao, Peng \& Hu}]{wang2019deep}
\bibinfo{author}{Wang, J.}, \bibinfo{author}{Chen, Y.}, \bibinfo{author}{Hao,
  S.}, \bibinfo{author}{Peng, X.}, \& \bibinfo{author}{Hu, L.}
  (\bibinfo{year}{2019}).
\newblock \bibinfo{title}{Deep learning for sensor-based activity recognition:
  A survey}.
\newblock {\it \bibinfo{journal}{Pattern Recognition Letters}\/},  {\it
  \bibinfo{volume}{119}\/}, \bibinfo{pages}{3--11}.
\bibitem[{Yang et~al.(2019)Yang, Jiang \& Guo}]{yang2019time}
\bibinfo{author}{Yang, C.}, \bibinfo{author}{Jiang, W.}, \&
  \bibinfo{author}{Guo, Z.} (\bibinfo{year}{2019}).
\newblock \bibinfo{title}{Time series data classification based on dual path
  cnn-rnn cascade network}.
\newblock {\it \bibinfo{journal}{IEEE Access}\/},  {\it \bibinfo{volume}{7}\/},
  \bibinfo{pages}{155304--155312}.
\bibitem[{Zhao et~al.(2017)Zhao, Lu, Chen, Liu \& Wu}]{zhao2017convolutional}
\bibinfo{author}{Zhao, B.}, \bibinfo{author}{Lu, H.}, \bibinfo{author}{Chen,
  S.}, \bibinfo{author}{Liu, J.}, \& \bibinfo{author}{Wu, D.}
  (\bibinfo{year}{2017}).
\newblock \bibinfo{title}{Convolutional neural networks for time series
  classification}.
\newblock {\it \bibinfo{journal}{Journal of Systems Engineering and
  Electronics}\/},  {\it \bibinfo{volume}{28}\/}, \bibinfo{pages}{162--169}.

\end{thebibliography}

\end{document}